%% file: ms.tex
\documentclass[letterpaper, 11pt]{proc}
\usepackage[utf8]{inputenc}
\usepackage{multicol}
\usepackage{graphicx}
\usepackage{amsmath}
\usepackage{float}
\usepackage{chngcntr}

\title{Quantitative Distortion Analysis of Flattening Applied to the Scroll from En-Gedi}
\author{C. Seth Parker\textdagger, W. Brent Seales\textdagger, Pnina Shor\textsection\\
{\footnotesize \textdagger\textit{Dept. of Computer Science, University of Kentucky, Lexington, KY, USA}}\\
{\footnotesize \textsection\textit{Israel Antiquities Authority, Jerusalem, Israel}}\\
{\footnotesize Correspondence to: c.seth.parker@uky.edu}
}
\date{}

\begin{document}

\input{000-titlepage}

\input{abstract}
\input{01-introduction}
\input{02-algorithms}
\input{03-metrics}
\input{04-results}
\input{05-conclusion}
\bibliographystyle{unsrt}
\bibliography{refs,seales}

\input{06-figures}
\end{document}

%% file: 000-titlepage.tex
\maketitle

%% file: abstract.tex
\section*{Abstract}
Non-invasive volumetric imaging can now capture the internal structure and detailed evidence of ink-based writing from within the confines of damaged and deteriorated manuscripts that cannot be physically opened. As demonstrated recently on the En-Gedi scroll, our “virtual unwrapping” software pipeline enables the recovery of substantial ink-based text from damaged artifacts at a quality high enough for serious critical textual analysis. However, the quality of the resulting images is defined by the subjective evaluation of scholars, and a choice of specific algorithms and parameters must be available at each stage in the pipeline in order to maximize the output quality.

A key stage in the virtual unwrapping pipeline is “flattening”, the process of transforming the 3D geometry of a manuscript’s wraps or pages into a 2D, flattened image suitable for reading and scholarship. Damage results in 3D geometries that are not necessarily isometric to a plane.  As a result, the computation of this transformation is nontrivial and may induce distortions. This distortion (stretching, shearing) is introduced into the 2D geometry, which in turn results in visual artifacts affecting the text visible in the flattened imagery. While many distortions are minor, any change in the text has the potential to affect subsequent scholarly analysis, such as paleography.

The computation of the flattening transformation is a crucial component of the “virtual unwrapping” pipeline.  Algorithms for flattening, including our physics-based material modeling algorithm, lead to questions about the tradeoffs in various approaches.  In this paper we provide a comparative analysis with particular emphasis on the visual distortions introduced by candidate flattening algorithms when applied to the problem of virtual unwrapping.  We show results from the unwrapped portions of the En-Gedi scroll.
%
%
%
%
%
%
%
%

%% file: 01-introduction.tex
\section{Introduction} \label{sec:intro}
The recent work of Seales \textit{et al.} demonstrates how non-invasive volumetric imaging and a ``virtual unwrapping'' software pipeline can recover substantial, ink-based text from within damaged manuscripts that cannot be physically opened \cite{seales-engedi-2016}. The En-Gedi scroll, excavated from the ruins of a Jewish synagogue on the western shores of the Dead Sea, was revealed to be a copy of the book of Leviticus, marking the scroll as one of the oldest known versions of the Pentateuchal book. The images generated by the virtual unwrapping pipeline were of such high quality that Segal \textit{et al.} were able to generate a full transcription and paleographic analysis of the text, describing them as being ``as readable as undamaged scrolls.'' \cite{segal-engedi-2016}
%
%

The flattening stage of the virtual unwrapping pipeline is particularly important to the quality of the output imagery. Preceding stages of the pipeline generate a three-dimensional geometric representation of the scroll's surface (a triangular mesh) and texture that surface with ink information extracted from the volumetric scan data. The 3D geometry of this surface is nonplanar, which makes viewing any text on its surface challenging. Flattening is the process of computing a 3D to 2D transformation (parameterization) that maps the mesh and associated texture information to a plane for ease of viewing.

\begin{figure}
  \centering
    \includegraphics[width=\columnwidth]{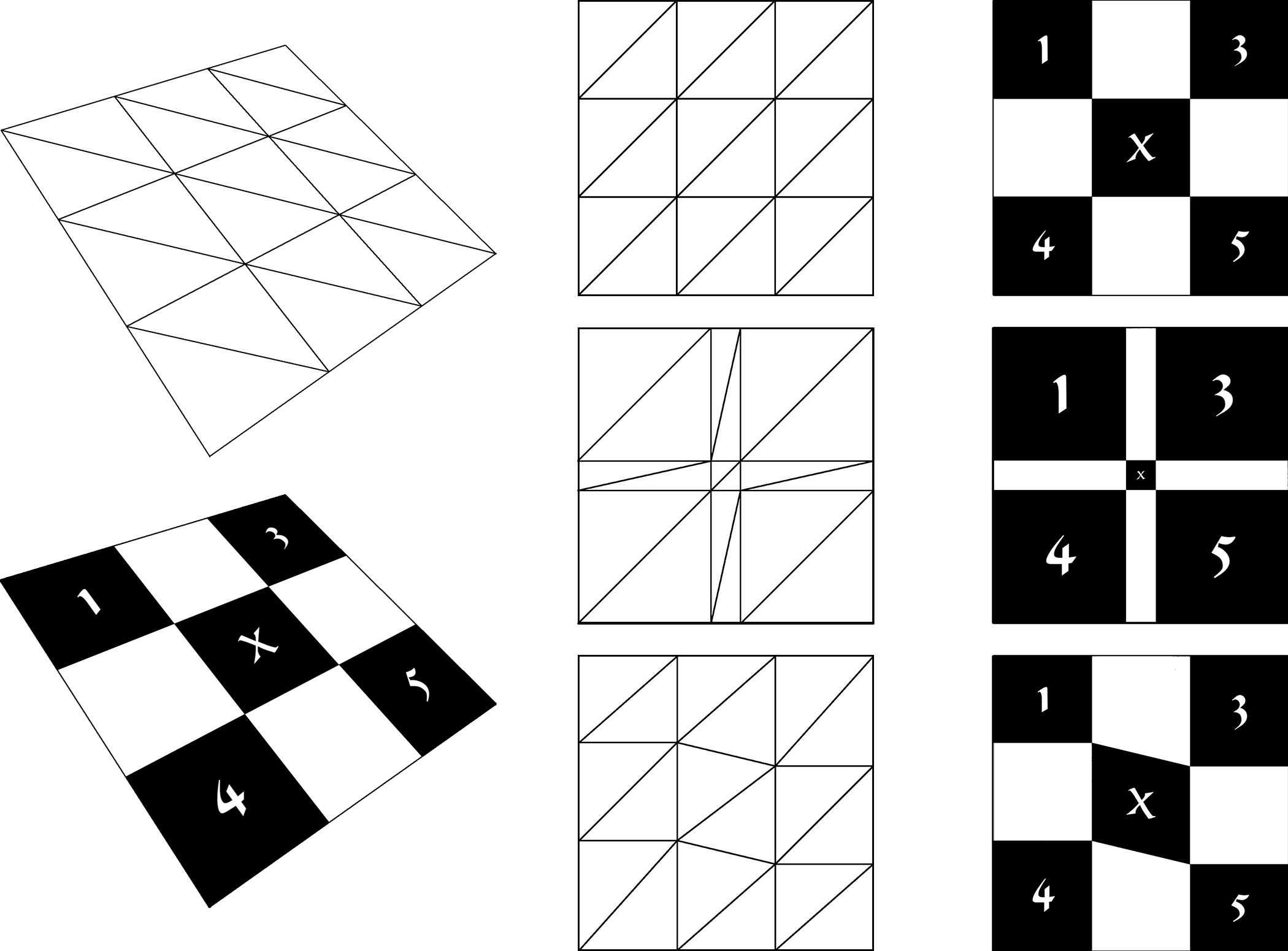}
    \caption{Comparison of area-maintaining (top), area-distorting (middle), and angle distorting (bottom) parameterizations of a textured 3D plane.}
    \label{fig:distortion}
\end{figure}

Surfaces are \textit{isometric} to a plane if, after being mapped to a plane, distances along the surface of the object are maintained. Similarly, surfaces that are \textit{conformal} to a plane maintain surface angles \cite{docarmo1976diffgeometry}. The highly irregular surfaces of damaged manuscript pages are not necessarily isometric or conformal. As a result, the computation of the flattening transformation is nontrivial and may induce distortion. This distortion is introduced into the 2D geometry, resulting in visual artifacts that affect the text seen in the flattened image. As shown in Figure \ref{fig:distortion}, a perfect parameterization results in a 2D image that maintains the relative scale and position of features in the output texture. When the area and angles of faces are not maintained, scaling and shearing artifacts can occur.

\begin{figure}
  \centering
    \includegraphics[width=\columnwidth]{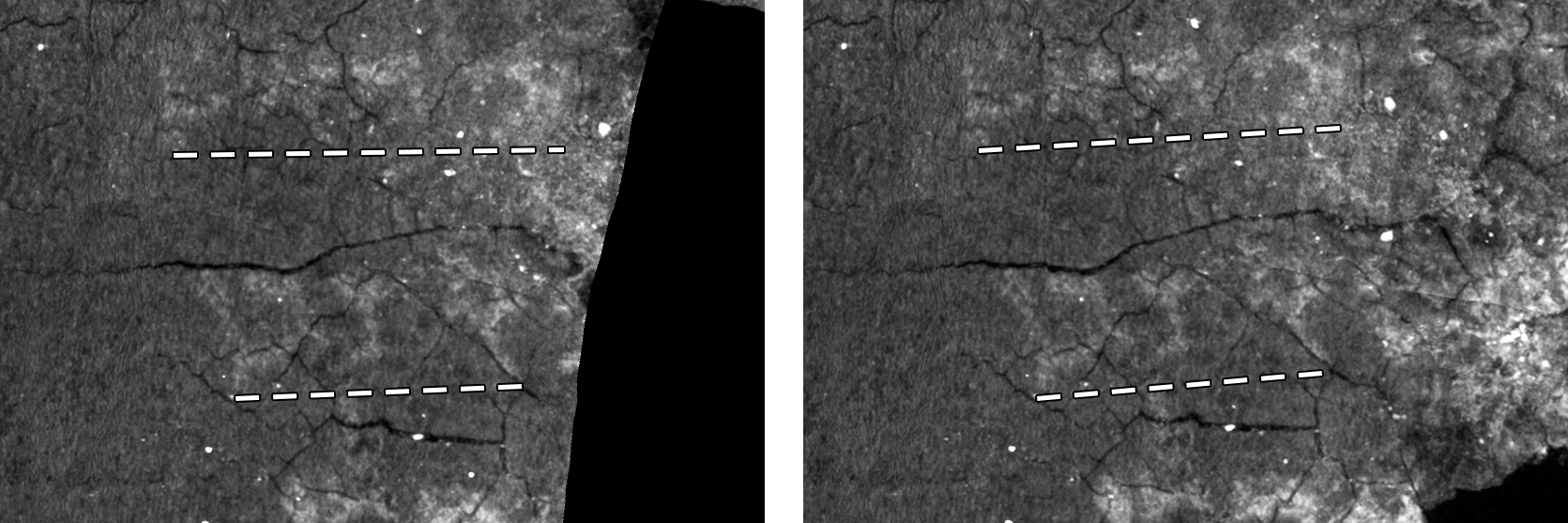}
    \caption{Example of mismatched surface features across overlapping texture images---Slight baseline shift.}
    \label{fig:baseline}
\end{figure}

Also of interest are the effects of parameterization when applied to the piecewise approximations of a much larger surface. Ideally, the virtual unwrapping pipeline would generate a single, continuous surface representation of the entire object being studied. In practice, however, the computation of this single surface is quite difficult, and therefore must be approached in a piecewise fashion. The result is a series of overlapping surface patches that must be merged. In lieu of a 3D merging technique, Seales \textit{et al.} performed a two-dimensional merge of each patch's flattened texture. In overlapping areas, distortion introduced by parameterization can lead to a visual mismatch in surface features. As shown in Figure \ref{fig:baseline}, a slight baseline shift occurs in the En-Gedi text when two overlapping texture images are merged.
%
%
%

While many distortions are minor, any change in the text has the potential to affect subsequent scholarly analysis. It is therefore essential that the distortion introduced by the flattening process be objectively measured and correlated to any resulting visual artifacts.

%% file: 02-algorithms.tex
\section{Methods}
Our goal is to measure and examine the effects of flattening algorithms when applied to seven surfaces extracted from the En-Gedi scroll. Each of these surfaces holds associated texture information generated by the texturing stage of our virtual unwrapping pipeline, and each must be flattened before a full reading can occur. To this end, we applied three flattening algorithms to the seven En-Gedi surfaces and used the resulting parameterizations to generate flattened texture images. The geometric error introduced by each flattening algorithm was then measured using the metrics outlined in Section \ref{sec:metrics} and correlated to the generated texture image. We pay particular attention to distortions introduced along the boundaries of the surfaces, locations that most affect the results of texturing merging.

\subsection{Flattening Algorithms} \label{sec:algorithms}
We compare two of the more popular geometric flattening algorithms---Least Squares Conformal Maps (LSCM)  \cite{levy2002lscm} and Angle-Based Flattening (ABF)  \cite{sheffer2001abf}. Both can be efficiently and quickly computed, produce parameterizations with low angular distortion, and preserve triangle orientations. We also examine our own physics-based material modeling (MM) algorithm that was used to flatten the En-Gedi scroll \cite{seales-engedi-2016}.

LSCM is a quasi-conformal parameterization technique that computes parameterized coordinates $(u,v)$ for each vertex $(x,y,z)$ such that the error of triangulation conformality is minimized in a least squares sense. It is a free-boundary technique that requires selecting only two vertices on the boundary of the mesh and fixing their parameterized coordinates. This makes it a good choice for very complicated surface geometries and boundaries. We use the LSCM implementation provided by libigl \cite{libigl2016}. 

ABF formulates the parameterization problem solely in terms of the measure of the interior angles of the triangulated mesh. The algorithm uses an iterative linear solver to minimize the relative error between the parameterized angles of the output space and a set of ``optimal'' angles computed from the original 3D angles of the mesh. Additional constraints are added to the equation to maintain the connectivity and validity of the resulting parameterized angles. Once the final angles have been computed, the position and magnitude of a single boundary edge is set in the parameterized space. The placement of this one edge, in combination with the parameterized angles, is enough to determine the $(u,v)$ coordinates of the other vertices in the mesh. For ABF, we make use of our modified implementation of the algorithm provided by Blender \cite{blender2016}.

Our own material modeling algorithm is designed specifically for flattening surfaces extracted from damaged textual objects. The algorithm assumes that the mesh being parameterized represents a physical object with its own set of physical properties. These properties--such as mass, density, and elasticity--are presumed to affect the results of the flattening process were it to be physically performed on the object. The MM method simulates these properties in the context of a mass-springs physics simulation, a common computer graphics technique for rendering cloth-like objects. While computationally expensive, this approach is highly configurable and allows for a more nuanced representation of the surface than its geometric properties might allow. The configuration of this system is custom, but is built on top of the Bullet Physics engine \cite{bullet2016}.

%% file: 03-metrics.tex
\subsection{Measuring Distortion} \label{sec:metrics}
A number of metrics already exist for quantifying the angular and area distortion introduced by parameterization. We use these metrics in conjunction with the generated texture images to guide our evaluation of the visual artifacts introduced by flattening distortion.

It is important to note that previous work has largely evaluated the effect of parameterization distortion as it pertains to mapping a 2D texture onto a 3D object. Parameterizations that are not isometric or conformal lead to texture information that is ``correct'' in 2D but which becomes distorted when mapped to 3D. In our case, texture information is derived from the 3D volume and then mapped to 2D space. Thus, parameterization error leads to 2D artifacts in texture information rather than 3D ones.

Sander \textit{et al.} define the $L^2$ and $L^{\infty}$ per-triangle texture stretch metrics \cite{sander2001texture}. These metrics describe the average and worst case stretch scenarios across the set of faces, respectively, and are an indicator of isometric error. $L^2$ is the root-mean-square stretch across the face of the considered triangle, while $L^{\infty}$ represents the largest local stretch---the greatest length obtained when mapping a unit vector from the parameterized triangle space to the 3D surface. We measure the corresponding per-mesh metrics as described by Sander \textit{et al.} and present the results in Tables \ref{tab:l2error} and \ref{tab:linferror}.

ABF introduces its own angular distortion metric for minimizing conformal error:
\begin{equation}
F(\alpha) = \sum_{i=1}^{T}\sum_{j=1}^{3} w_j^i(\alpha_j^i - \phi_j^i)^2
\end{equation}
where $i$ iterates through the set of face triangles $T$, $j$ iterates through the angles of each face, $w_j^i$ is a per-angle weight, and $\alpha_j^i$ and $\phi_j^i$ are the parameterized and optimal angles respectively. This is the same metric used as part of the minimization task and can be evaluated for the whole mesh by averaging the error across the total number of angles:

\begin{equation}
F(M)=\frac{F(\alpha)}{3T}
\end{equation}

Blender, the open-source 3D modeling and rendering application, utilizes both angular and area error metrics for displaying parameterization error to the user \cite{blender2016}. We make use of the per-triangle area metric:
\begin{equation}
\begin{gathered}
\alpha = \frac{A(T)}{A(M)},\ \beta = \frac{A(t)}{A(m)}\\
E(t) = 
\begin{cases}
      1.0-\frac{\beta}{\alpha},& \text{if } \alpha>\beta\\
      1.0-\frac{\alpha}{\beta},& \text{otherwise}
\end{cases}
\end{gathered}
\end{equation}
where $A(x)$ is the area function, $T$ and $M$ represent a 3D triangle and mesh, and $t$ and $m$ represent the corresponding 2D triangle and mesh. We calculate a per-mesh value $E(M)$ by taking the mean of the per-triangle error. We also made use of Blender's parameterization error visualization to produce the error distribution images shown in Appendix \ref{sec:figures}.

%% file: 04-results.tex
\section{Results} \label{sec:results}
The observed error metrics and the flattened textures for each of the En-Gedi meshes (EG0-3,5,6) are shown in Appendix \ref{sec:figures}. ABF and LSCM both failed to produce a parameterization for EG4, thus we only present the metrics for its MM results.

With a few exceptions, all of the parameterization algorithms perform similarly in terms of angular distortion (Table \ref{tab:alpherror}). Even for meshes that appear to demonstrate significantly varying error values between ABF, LSCM, and MM, such as with mesh EG1, the visual differences on a global scale appear to be minor (Figure \ref{fig:eg1-params}). Figure \ref{fig:eg0-angleerror} visualizes a typical distribution pattern for the angular error across the parameterized surface. Since ABF and LSCM both seek to minimize conformal error (angular distortion) this is not surprising. That MM produces similar results is notable since it makes no explicit guarantees for minimizing conformal error.

Much of the error and resulting visual distortion is represented by the isometric error metrics. MM produced significantly worse $L^2$ and $L^\infty$ metric values, which resulted in artifacts such as the slight global shear and elongation of the right edge of the MM image for EG0 (Figure \ref{fig:eg0-params}). This distortion is better visualized in Figure \ref{fig:eg0-areaerror}, where areas of higher area distortion streak the right side of the parameterization. While these isolated areas demonstrated large amounts of stretch, the effect on global area distortion was minimal (Table \ref{tab:blenderror}). LSCM parameterizations, on the other hand, performed similarly to ABF for $L^2$ stretch, but much worse on area distortion. In some cases, this resulted in very noticeable nonlinear feature scaling in the texture images, as seen in Figures \ref{fig:eg2-params} and \ref{fig:eg6-params}. 

\begin{figure}
\centering
    \includegraphics[width=\columnwidth]{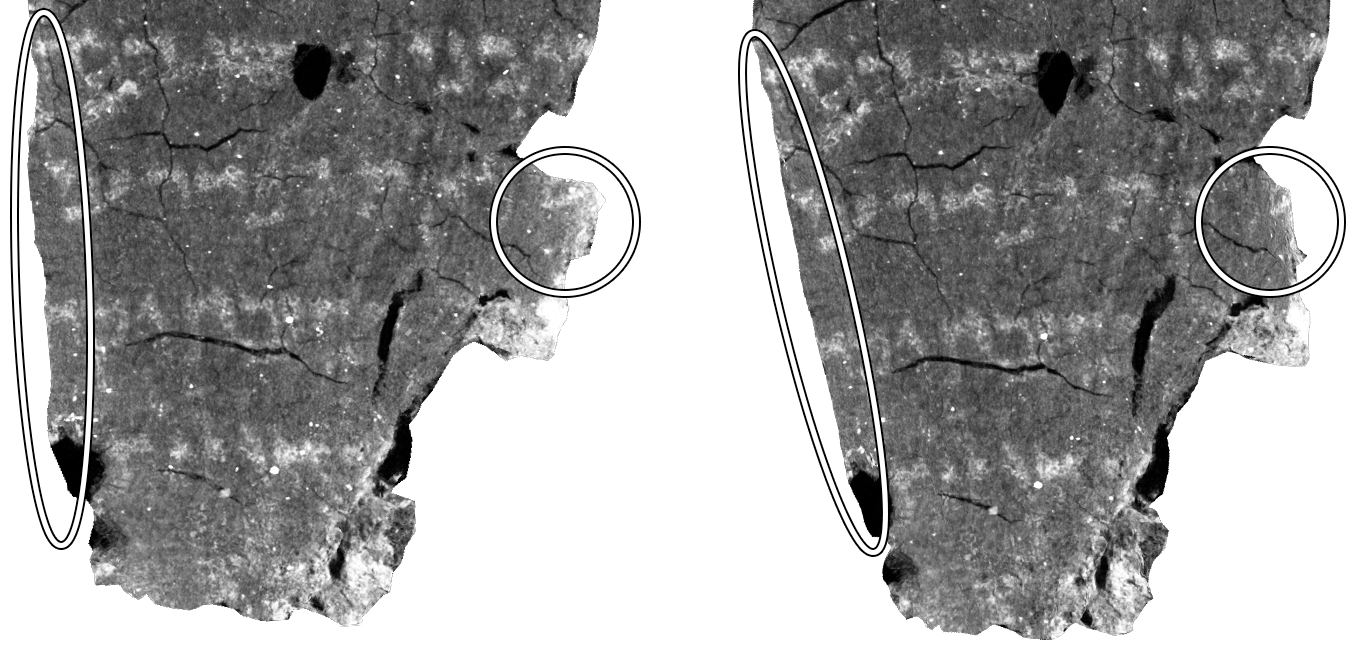}
    \caption{Differences in EG2 parameterizations: ABF (left) and MM (right).}
    \label{fig:minordiffs}
\end{figure}

Also worth mentioning are some distortions produced by MM that are not easily represented by the considered metrics. As shown in Figure \ref{fig:minordiffs}, the internal features of the ABF and MM texture images are largely consistent, but the outer borders demonstrate significant differences. The MM image displays a smoother left edge and a ``missing'' boundary feature on the right edge. The absence of this feature was caused by a group of adjacent faces that did not completely flatten, but instead ``folded'' onto each other in parameter space. Unlike ABF, MM in its current form makes no guarantees about the shape of the mesh border, thus allowing for edge cases such as these.

%% file: 05-conclusion.tex
\section{Conclusion}
We can conclude that for many applications, Angle-based Flattening is the preferred first choice for surface flattening algorithms. It requires little to no configuration and demonstrates low isometric and conformal error with few strong visual artifacts. Future work should focus on implementing the extension to Angled-based Flattening, ABF++ \cite{sheffer2005abf++}. ABF++ modifies the minimized error formulation to improve computation speeds for meshes with large numbers of vertices. As the virtual unwrapping pipeline continues to improve and surface meshes begin to represent larger and larger areas, these sorts of performance considerations will become extremely important.

Our own material modeling algorithm showed angular distortion similar to that of ABF, but demonstrated problematic stretching artifacts throughout the surface. Many of these artifacts could likely be corrected if the MM system was configured to model the physical properties of the En-Gedi scroll with extreme accuracy. While this could potentially produce better flattening results than ABF, the benefits would likely be outweighed by the large amount of time spent configuring the system. Fortunately, many of the current distortions are isolated to small areas of the meshes and do not produce large scale visual artifacts.

The importance of quantifying and visualizing error introduced by flattening cannot be understated. When flattening surfaces that are neither isometric nor conformal to a plane, the introduction of visual artifacts into the resulting flattened image is inevitable. These artifacts are often subtle and evenly distributed across the entire output image, making it difficult to evaluate the quality of a virtually unwrapped surface. By using geometric measures of error to guide our process, we were able to objectively measure distortion in the En-Gedi surfaces and correlate that distortion to specific visual artifacts in the flattened texture images, making qualitative assessment much more direct. 

\section{Acknowledgments}
C.S.P. acknowledges the invaluable professional contributions of C. Chapman in the editorial preparation of this manuscript. We thank C. Gardella for her contributions to data preparation and visualization. Thanks also to the Dead Sea Scrolls Unit of the IAA team for enabling this project. W.B.S. acknowledges funding from the NSF (awards IIS-0535003 and IIS- 1422039). Any opinions, findings, and conclusions or recommendations expressed in this material are those of the author(s) and do not necessarily reflect the views of the NSF. W.B.S. acknowledges funding from Google and support from S. Crossan (Founding Director of the Google Cultural Institute).

%% file: 06-figures.tex
\appendix
\counterwithin{figure}{section}
\counterwithin{table}{section}
\section{Figures and Tables} \label{sec:figures}

\setlength{\tabcolsep}{10pt}
\renewcommand{\arraystretch}{1.25}

\begin{table}[!htb]
\begin{center}
\caption{$L^2$ Error} \label{tab:l2error}
\begin{tabular}{ c || c | c | c }
Mesh & ABF & LSCM & MM\\
\hline
EG0 & 1.00340 & 1.03717 & 2.45248\\
EG1 & 1.00237 & 1.00790 & 2.13652\\
EG2 & 1.00896 & 1.13735 & 2.30123\\
EG3 & 1.00261 & 1.03728 & 4.76168\\
EG4 & --- & --- & 7.46261\\
EG5 & 1.05035 & 1.11776 & 23.4113\\
EG6 & 1.00375 & 1.27112 & 1.65306\\
\end{tabular}
\end{center}
\end{table}

\begin{table}[!htb]
\begin{center}
\caption{$L^\infty$ Error} \label{tab:linferror}
\begin{tabular}{ c || c | c | c }
Mesh & ABF & LSCM & MM\\
\hline
EG0 & 1.29810 & 1.85130 & 356.085\\
EG1 & 36.58980 & 4.41805 & 532.838\\
EG2 & 14.32470 & 24.67560 & 548.357\\
EG3 & 5.96818 & 22.69540 & 882.906\\
EG4 & --- & --- & 1957.460\\
EG5 & 40.62920 & 32.87700 & 8532.760\\
EG6 & 2.37835 & 11.15980 & 308.242\\
\end{tabular}
\end{center}
\end{table}

\begin{table}[!htb]
\begin{center}
\caption{$F(M)$ Error} \label{tab:alpherror}
\begin{tabular}{ c || c | c | c }
Mesh & ABF & LSCM & MM\\
\hline
EG0 & 0.63695 & 0.63700 & 0.63574\\
EG1 & 12.26375 & 9.03262 & 8.96267\\
EG2 & 0.90651 & 0.89886 & 0.90160\\
EG3 & 2.29791 & 2.26831 & 2.26686\\
EG4 & --- & --- & 2.29255\\
EG5 & 0.54397 & 0.54145 & 0.54505\\
EG6 & 0.37898 & 0.38032 & 0.37896\\
\end{tabular}
\end{center}
\end{table}

\begin{table}[!tb]
\begin{center}
\caption{$E(M)$ Error} \label{tab:blenderror}
\begin{tabular}{ c || c | c | c }
Mesh & ABF & LSCM & MM\\
\hline
EG0 & 0.28029 & 0.32231 & 0.28775\\
EG1 & 0.28197 & 0.29037 & 0.28410\\
EG2 & 0.28746 & 0.39442 & 0.28937\\
EG3 & 0.28434 & 0.31817 & 0.29602\\
EG4 & --- & --- & 0.30060\\
EG5 & 0.29598 & 0.35874 & 0.30059\\
EG6 & 0.28175 & 0.42408 & 0.28412\\
\end{tabular}
\end{center}
\end{table}

\onecolumn
\begin{figure}[!htb]
\centering
    \includegraphics[height=0.40\textheight]{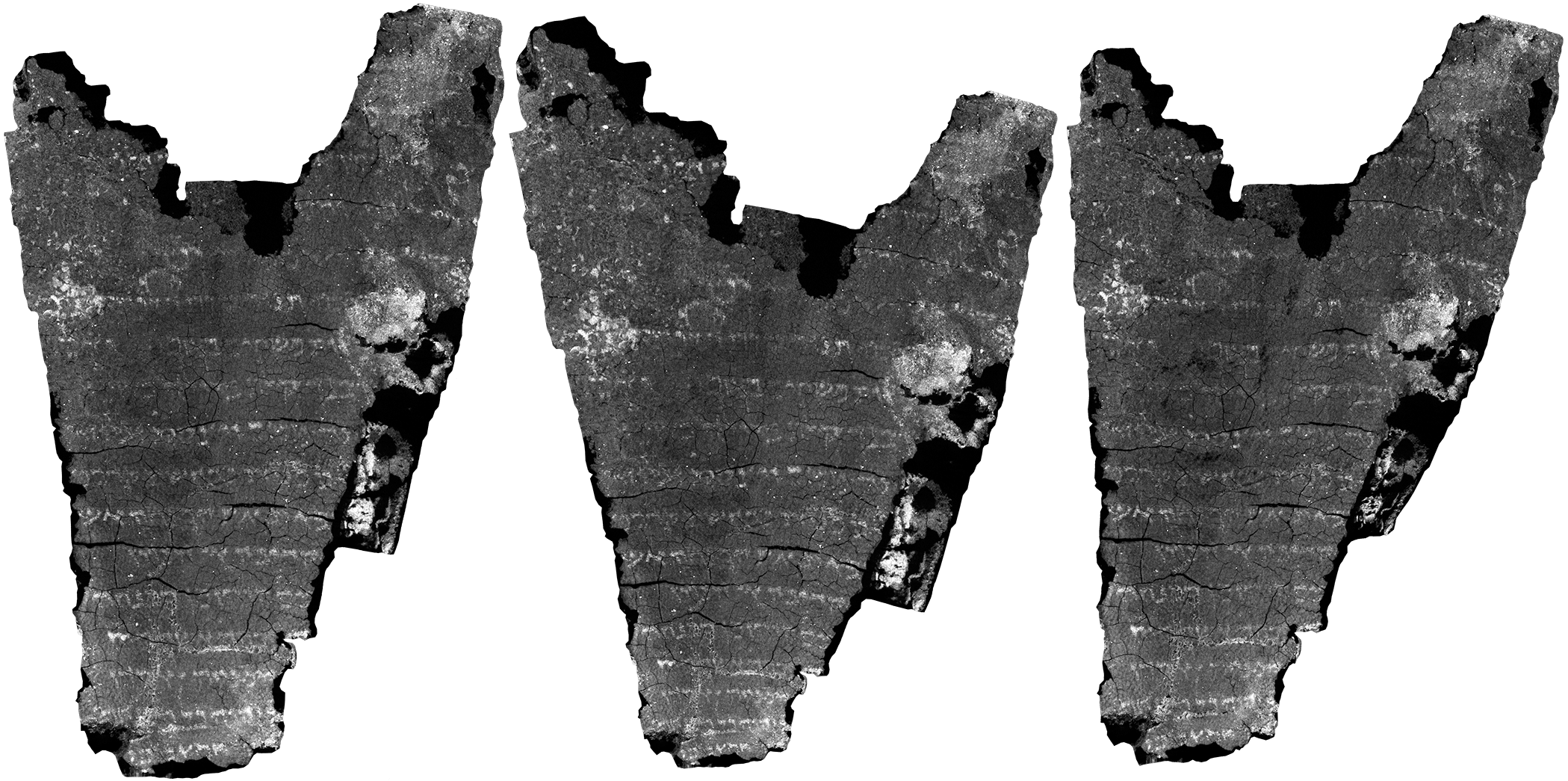}
    \caption{EG0 parameterizations: ABF (left), LSCM (middle), MM (right)}
    \label{fig:eg0-params}
\end{figure}

\begin{figure}[!htb]
\centering
    \includegraphics[height=0.40\textheight]{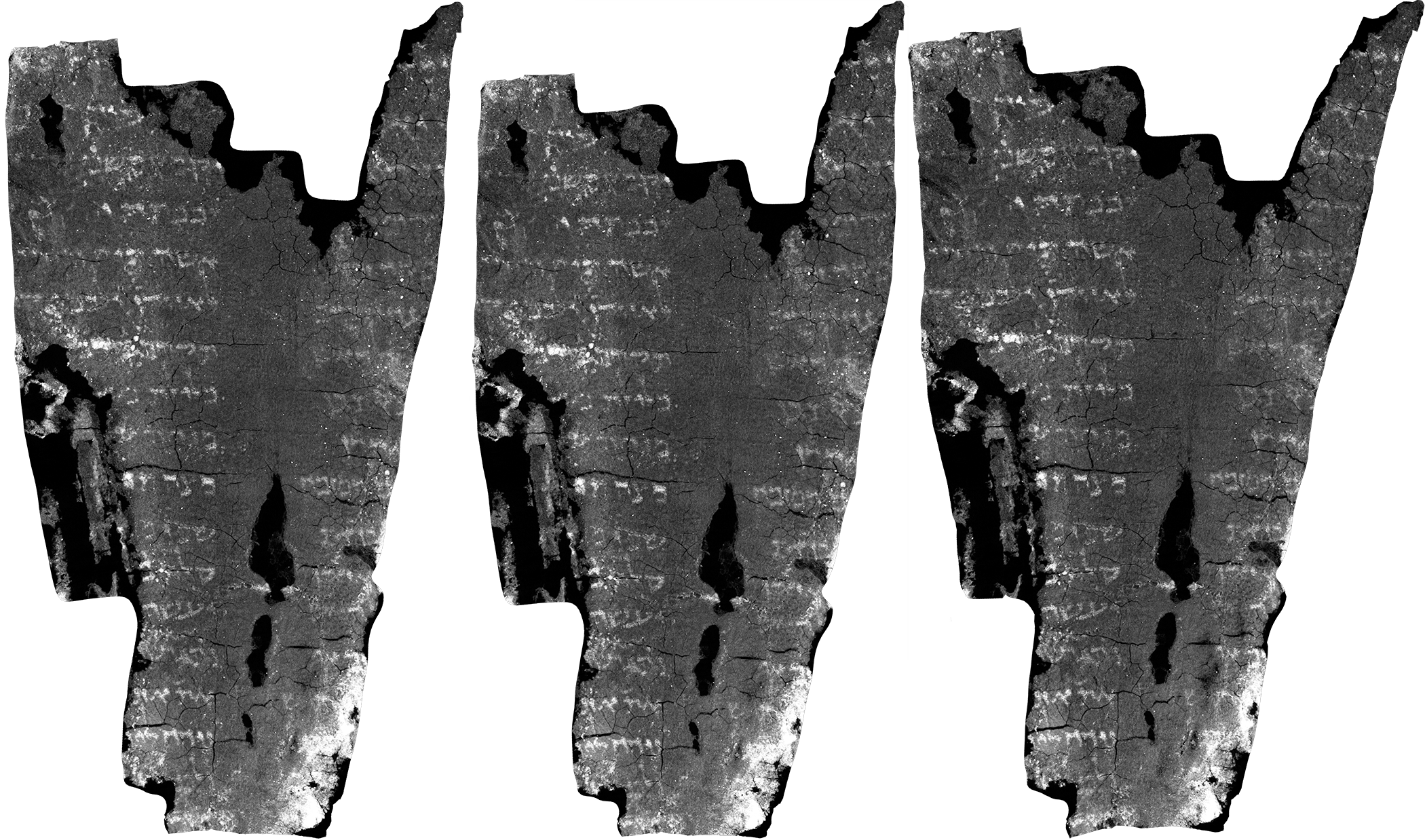}
    \caption{EG1 parameterizations: ABF (left), LSCM (middle), MM (right)}
    \label{fig:eg1-params}
\end{figure}

\begin{figure}[!htb]
\centering
    \includegraphics[height=0.40\textheight]{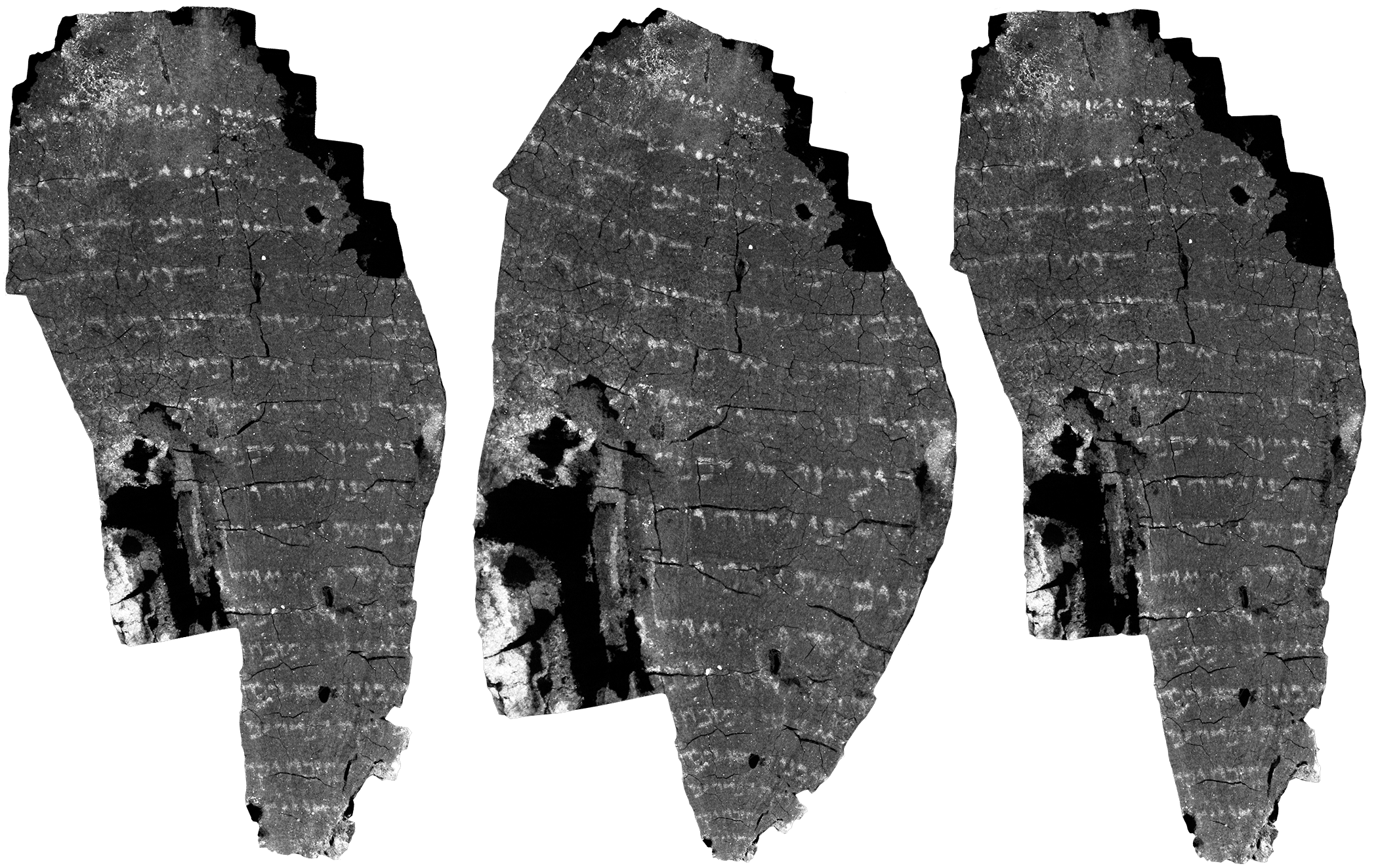}
    \caption{EG2 parameterizations: ABF (left), LSCM (middle), MM (right)}
    \label{fig:eg2-params}
\end{figure}

\begin{figure}[!htb]
\centering
    \includegraphics[height=0.40\textheight]{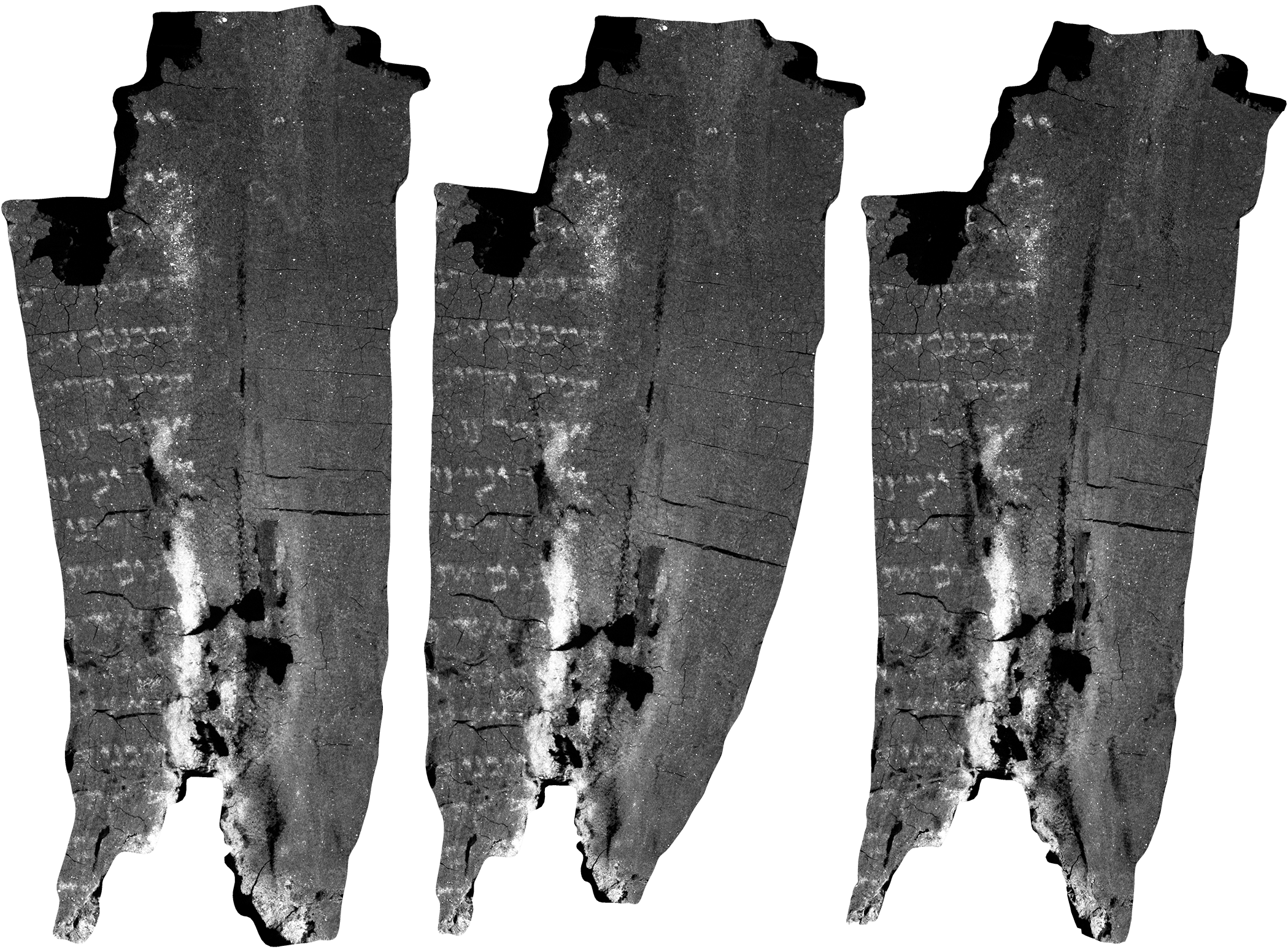}
    \caption{EG3 parameterizations: ABF (left), LSCM (middle), MM (right)}
    \label{fig:eg3-params}
\end{figure}

\begin{figure}[!htb]
\centering
    \includegraphics[height=0.40\textheight]{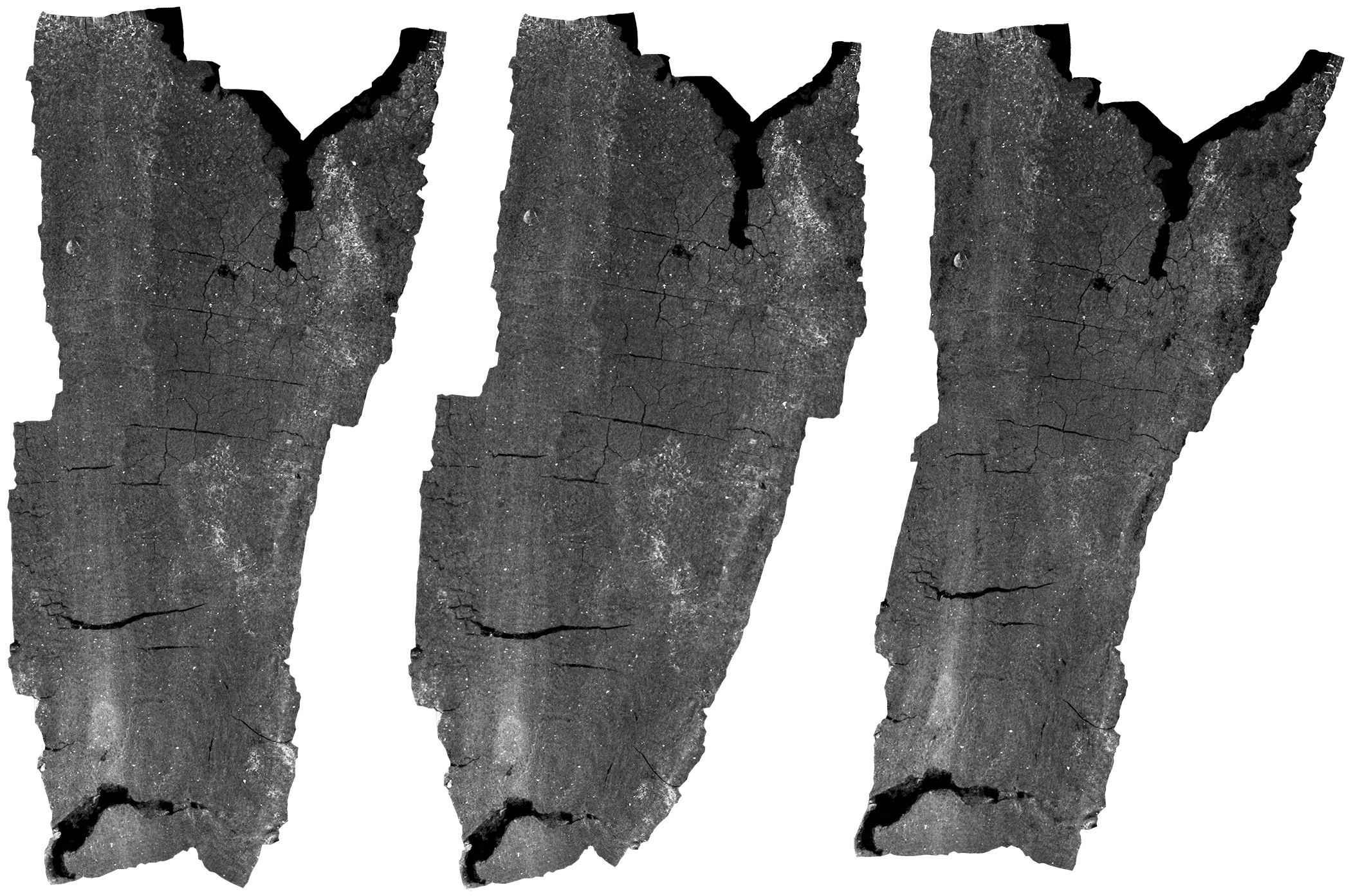}
    \caption{EG5 parameterizations: ABF (left), LSCM (middle), MM (right)}
    \label{fig:eg5-params}
\end{figure}

\begin{figure}[!htb]
\centering
    \includegraphics[height=0.40\textheight]{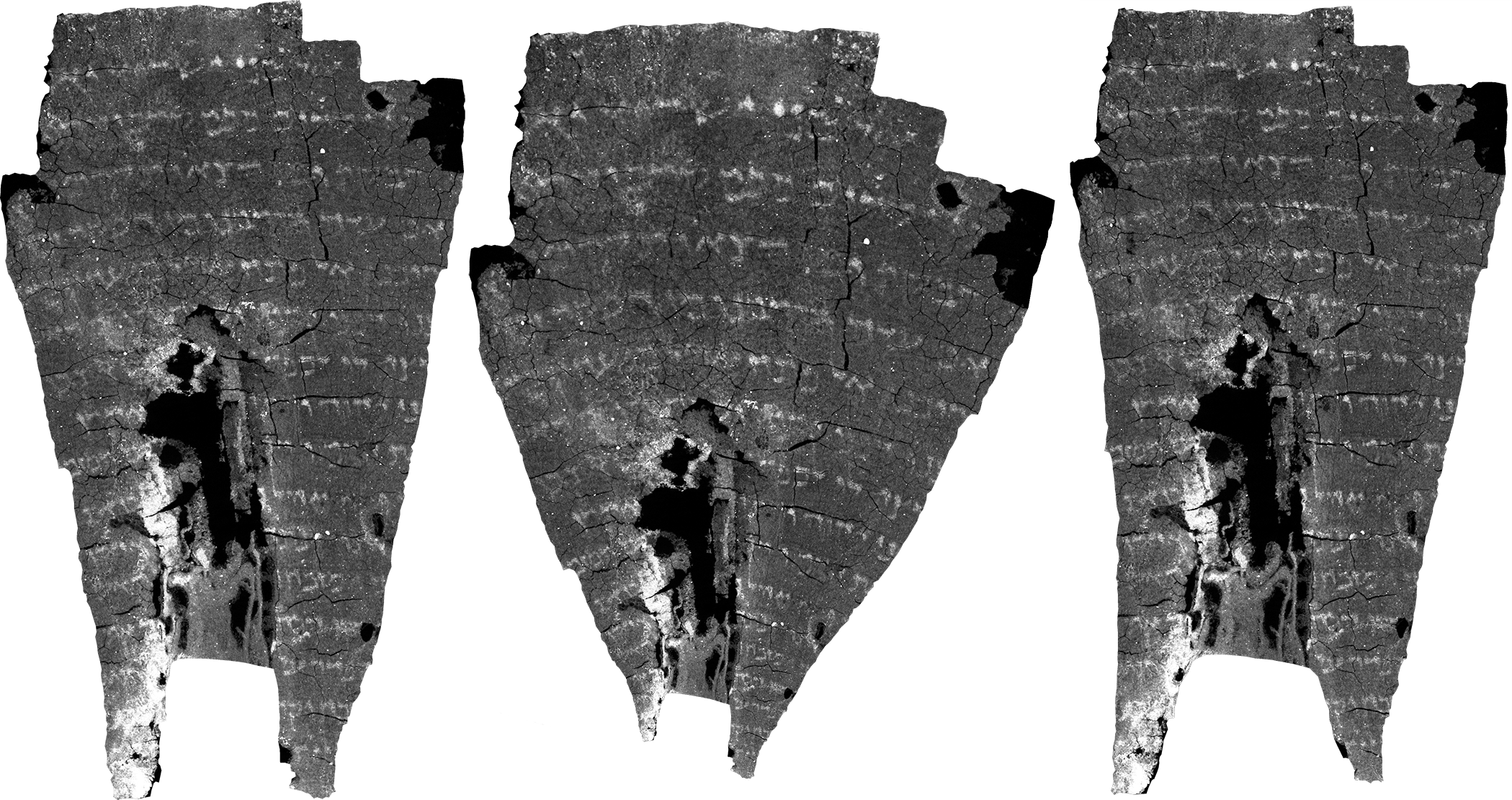}
    \caption{EG6 parameterizations: ABF (left), LSCM (middle), MM (right)}
    \label{fig:eg6-params}
\end{figure}

\begin{figure}[!htb]
\centering
    \includegraphics[height=0.40\textheight]{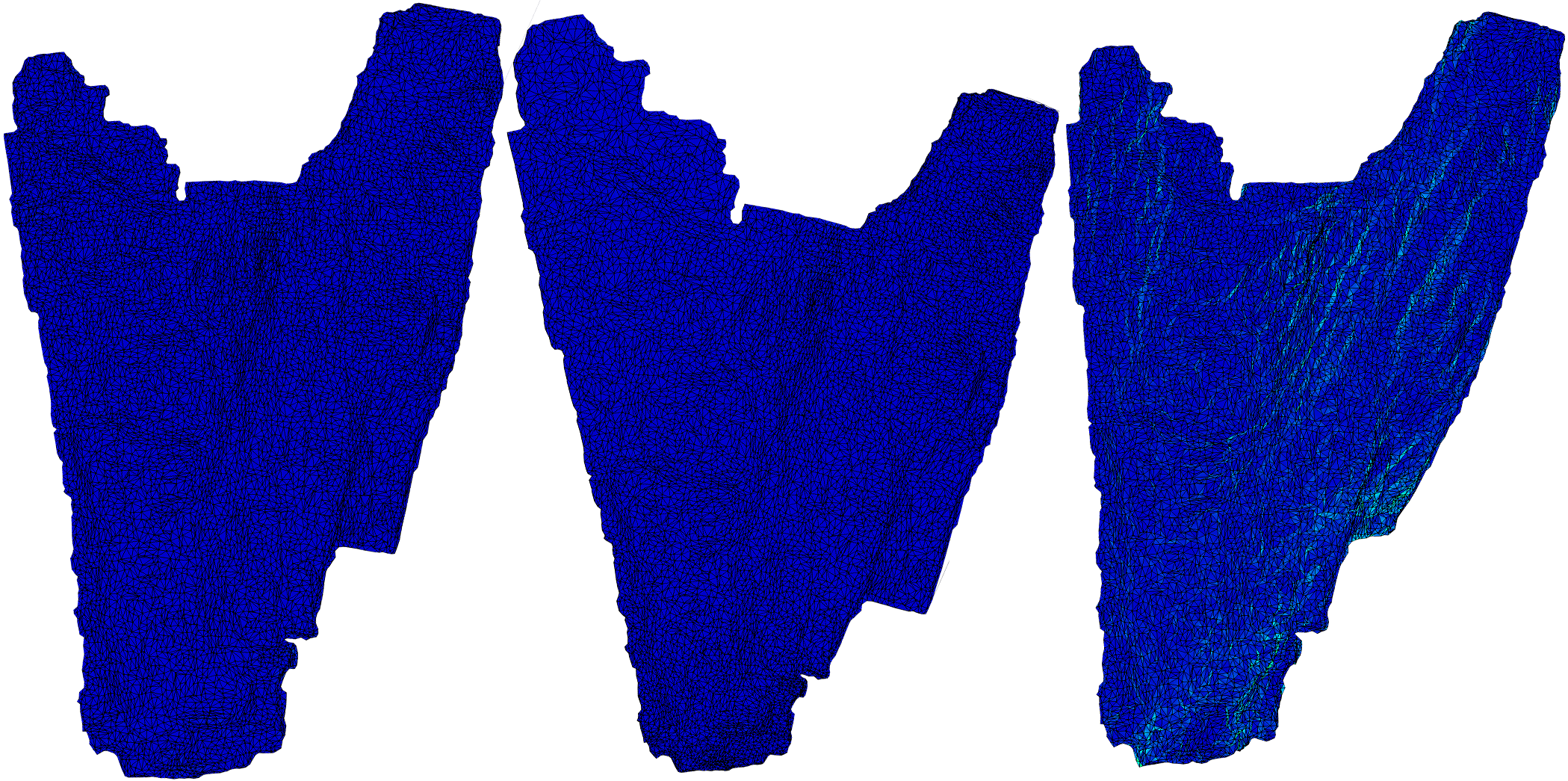}
    \caption{Distribution of Angular Error for EG0: ABF (left), LSCM (middle), MM (right)}
    \label{fig:eg0-angleerror}
\end{figure}

\begin{figure}[!htb]
\centering
    \includegraphics[height=0.40\textheight]{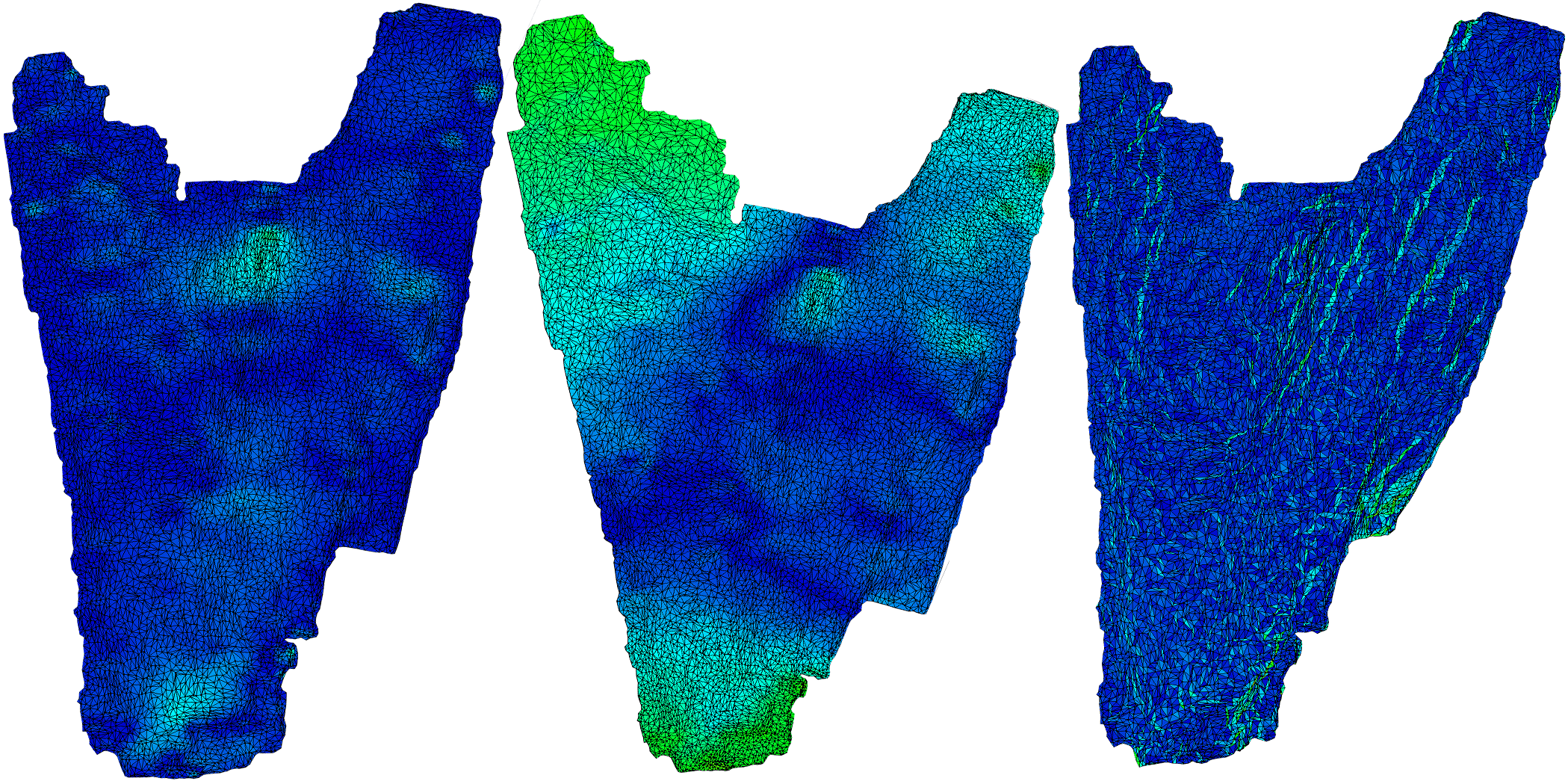}
    \caption{Distribution of Area Error for EG0: ABF (left), LSCM (middle), MM (right)}
    \label{fig:eg0-areaerror}
\end{figure}